\RequirePackage{fix-cm}
\documentclass[smallextended]{svjour3}       
\RequirePackage{amsmath}

\smartqed  
\usepackage{graphicx}
\usepackage[numbers]{natbib} 
\usepackage{color}
\usepackage[colorlinks]{hyperref}
\usepackage{enumitem}
\usepackage{tabularx,longtable,multirow,caption}
\usepackage[style=base]{subcaption}
\begin{document}

\title{Estimation of Tissue Oxygen Saturation from RGB images and Sparse Hyperspectral Signals based on Conditional Generative Adversarial Network \thanks{} }


\author{Qingbiao Li$^{1,2}$         \and
        Jianyu Lin$^{1,3}$  \and
        Neil T. Clancy$^{4,5}$  \and
        Daniel S. Elson$^{1,2}$  
}


\institute{Qingbiao Li 
            \email{ql716@ic.ac.uk} \\ 
           Jianyu Lin    
            \email{xjtuljy@gmail.com}\\  
            Neil T.Clancy  
            \email{n.clancy@ucl.ac.uk}\\
            Daniel S. Elson   
            \email{daniel.elson@imperial.ac.uk}\\
            1 The Hamlyn Centre for Robotic Surgery, Imperial College London, London, UK \\
            2 Department of Surgery and Cancer, Imperial College London, London, UK \\
            3 Department of Computing, Imperial College London, London, UK \\
            4 Wellcome/EPSRC Centre for Interventional and Surgical Sciences (WEISS), University College London, UK \\
            5 Centre for Medical Image Computing (CMIC), Department of Computer Science, University College London, UK
}

\date{Received: date / Accepted: date}

\maketitle

\begin{abstract}
\mbox{}\\
\textcolor{black}{\textbf{\textit{Purpose}} Intra-operative measurement of tissue oxygen saturation ($StO_2$) is important in detection of ischaemia, monitoring perfusion and identifying disease. Hyperspectral imaging (HSI) measures the optical reflectance spectrum of the tissue and uses this information to quantify its composition, including $StO_2$. However, real-time monitoring is difficult due to capture rate and data processing time.}\\
\textcolor{black}{\textbf{\textit{Methods}} An endoscopic system based on a multi-fibre probe was previously developed to sparsely capture HSI data (sHSI). These were combined with RGB images, via a deep neural network, to generate high resolution hypercubes and calculate $StO_2$. To improve accuracy and processing speed, we propose a dual-input conditional generative adversarial network (cGAN), Dual2StO2, to directly estimate $StO_2$ by fusing features from both RGB and sHSI.}
\\
\textcolor{black}{\textbf{\textit{Results}} Validation experiments were carried out on \textit{in vivo} porcine bowel data, where the ground truth $StO_2$ was generated from the HSI camera. Performance was also compared to our previous super-spectral-resolution network, SSRNet in terms of mean $StO_2$ prediction accuracy and structural similarity metrics. Dual2StO2 was also tested using simulated probe data with varying fibre number.}
\\
\textcolor{black}{\textbf{\textit{Conclusions}} {$StO_2$ estimation by Dual2StO2 is visually closer to ground truth in general structure, achieves higher prediction accuracy and faster processing speed than SSRNet. Simulations showed that results improved when a greater number of fibres are used in the probe. Future work will include refinement of the network architecture, hardware optimisation based on simulation results, and evaluation of the technique in clinical applications beyond $StO_2$ estimation.}}

\vspace{-1.0em}
\end{abstract}
\vspace{-1.0em}
\section{Introduction} \label{intro}
\vspace{-1.0em}
\textcolor{black}{Tissue perfusion and oxygenation are important clinical indicators of organ health during minimal access surgery (MAS). Endoscopic hyperspectral imaging (HSI) is a non-invasive optical technique to capture quantitative spectral information with a high spatial resolution based on narrow spectral bands over a virtually continuous spectral range for live tissue diagnostics and monitoring \cite{Lu2014}. HSI can  be used to estimate oxygen saturation ($StO_2$) and perfusion, which reflects tissue function and the health of organ's blood supply. This, in turn, can be applied to various important clinical applications  \cite{Lu2014}, including monitoring of cortical haemodynamics during brain surgery \cite{mori2014}, reperfusion during organ transplantation \cite{Clancy2016} and detection of intestinal ischaemia \cite{akbari2010}. High resolution spectral data can also be used to characterise tissue and detect subtle differences between normal and dysplastic areas \cite{kumashiro2016}. HSI is a non-contact technique, compatible with conventional surgical light sources and endoscopes, and has some important advantages over competing optical techniques, such as photoacoustic tomography (PAT) \cite{wang2012photoacoustic}, which requires ultrasound contact and a complex laser source.}

\textcolor{black}{HSI requires acquisition of a hypercube, which has one spectral and two spatial dimensions. Imaging hardware may use tuneable filters or spatial scanning, but does not typically achieve real-time operation due to the data capture and processing times. Snapshot spectral imaging acquires the entire hypercube simultaneously, but the number of wavelengths or spatial resolution must be sacrificed to achieve high speed acquisition. This trade-off between spectral information, spatial resolution and acquisition speed is a barrier for clinical use of HSI and other optical imaging techniques \cite{Lu2014}}. 

\begin{figure}[!ht]
	\centering
	\begin{subfigure}[b]{0.41\textwidth}
		\centering
		\includegraphics[width=\textwidth]{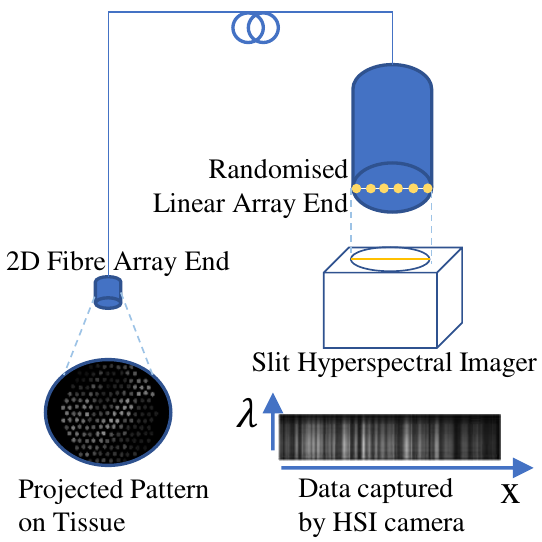}
		\caption{The spatial mapping}
		\label{fig:Spatial_mapping_bundle}
	\end{subfigure}
	\begin{subfigure}[b]{0.58\textwidth}
		\centering
		\includegraphics[width=0.9\textwidth]{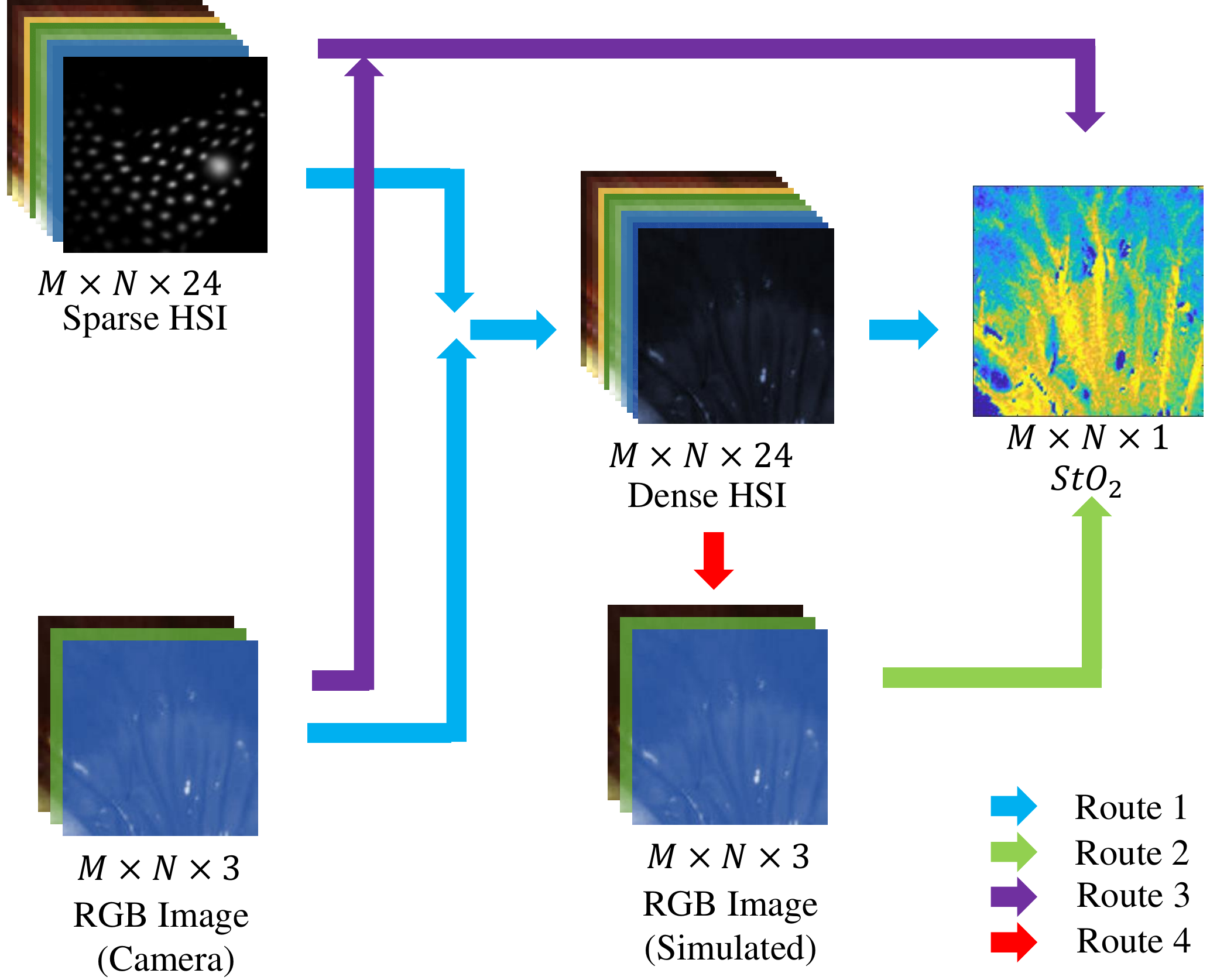}
		\caption{Route}
		\label{fig:Route}
	\end{subfigure}
	\caption{a: The spatial mapping between randomised end (2D) and the linear end (1D); b: Processing methods for $StO_{2}$ estimation.}
    \vspace{-1.5em}
\end{figure}

\textcolor{black}{To overcome this}, we \textcolor{black}{previously} developed a dual mode structured light and hyperspectral imaging (SLHSI) system  \cite{clancy2011}\cite{lin2018} to capture sparse hyperspectral images in real-time, \textcolor{black}{as illustrated in Fig.~\ref{fig:Spatial_mapping_bundle}. The light (i.e. reflectance or fluorescence) from the tissue surface was imaged onto the 2D fibre array, and the bundle randomly re-ordered the fibres into a linear array at the other end. The spectrum carried by each fibre could then be captured by imaging the linear array onto  the entrance slit of an imaging spectrograph. The data could then be rearranged computationally to generate sparse hyperspectral images (sHSI) in a snapshot. The 2.8 mm fibre bundle can be inserted through an endoscope biopsy port or attached to the endoscope or another surgical instrument \cite{clancy2011}. The system could also be used to record spectrally-encoded structured lighting (SL) images \cite{clancy2011}\cite{lin2018}, although this capability is not explored further in this paper.}

\textcolor{black}{To process the acquisition, a} super-spectral-resolution network, called SSRNet, was proposed to integrate dense RGB images and \textcolor{black}{sHSI} for pixel-level hypercube estimation \cite{lin2018}. The hypercube could be used to estimate $StO_2$ based on the modified Beer-Lambert law as illustrated in processing Route 1 (Blue line) in Fig.~\ref{fig:Route}. \textcolor{black}{Previous work also explored the} feasibility of estimating $StO_2$ directly from RGB images (Route 2, Fig.~\ref{fig:Route}) \cite{li2018estimation}, and showed that hyperspectral information improves the accuracy of  the result \cite{Lin2017}. However, as the aim of SSRNet was to predict \textcolor{black}{dense HSI} hypercubes, it was not explicitly optimised for $StO_2$ estimation, and the value of combining RGB images with sparse \textcolor{black}{HSI} hypercubes has not been evaluated for estimating dense $StO_2$ maps.  

\textcolor{black}{In this paper, we extend the previous published results, \textcolor{black}{proposing} a dual input network using cGAN, Dual2StO2, to achieve dense $StO_2$ estimation using end-to-end learning, without the need for the intermediate spectral estimation step. The proposed network was inspired by the performance of GANs in super-resolution \cite{ledig2017photo}\cite{yi2018generative},  to achieve super-resolution estimation in spatial (for sHSI) and spectral (for RGB) domains. A minimax two-player game was utilised between the generator and discriminator to further improve the accuracy of per-pixel regression for $StO_2$ estimation. By adding conditional input, the generator in cGAN would estimate StO2 imitating the structure of the condition, instead of random image generation in GANs. The results from Mirza \textit{et al.} \cite{mirza2014} and Isola \textit{et al.} \cite{pix2pix2016} also supported that cGAN could achieve higher pixel-level accuracy than other GANs with the same settings.} 
\textcolor{black}{The relationship between two input modalities (RGB, sHSI) and output (StO2) were known a priori, which enabled the network to be trained by supervised learning, achieving faster convergence and prediction accuracy. Additionally, a customized mask was added to filter saturated pixels and unreliable estimates at the pixel level.}
\textcolor{black}{Furthermore, one of the key parameters in designing the MSI data acquisition system is the number of fibres in the bundle, and we have therefore additionally simulated the performance of this estimation for different fibre bundles. This approach is represented by Route 3 (Purple line) in Fig.~\ref{fig:Route}. This will allow optimisation of future hardware designs to increase robustness.}\\
In this paper, Sec.~\ref{sec:Data_And_Preprocessing} will describe data acquisition and HSI data synthesis, while Dual2StO2 is presented in Sec.~\ref{sec:Dual2StO2}. The evaluation metrics and validation setup for this method are described in Sec.~\ref{sec:exp_and_result}, followed by a validation of the network via an inter-\textcolor{black}{animal} study on porcine bowel \textit{in vivo}. The previous two-stage $StO_2$ estimation approach (Route 1, Blue line) developed by Lin \textit{et al.} \cite{Lin2017} in Fig.~\ref{fig:Route} was adopted as the baseline against which the performance of the proposed network was evaluated.
\vspace{-1.0em}
\section{Materials and methods}
\vspace{-1.0em}
\subsection{Data Acquisition and Preprocessing} \label{sec:Data_And_Preprocessing}
\vspace{-1.0em}
The porcine bowel \textit{in vivo} data was captured by a liquid crystal tuneable filter (LCTF) in the range of $460 - 700 \,\mathrm{nm}$ with $10 \,\mathrm{nm}$ interval, as described in a previous work \cite{Clancy2015}. Here, a subset of the spectral data from $460 - 690 \,\mathrm{nm}$ was considered as a ground truth 24-channel hypercube with spatial size $256 \times 192$ pixels. A total of 50 acquisitions were selected from 15 separate animals.
\noindent \textbf{Simulated RGB images.}
\textcolor{black}{The RGB image (Input - x) was simulated from the hypercube (Route 3 in Fig.~\ref{fig:Route}) using the known spectral response of a colour camera \cite{Clancy2015} \cite{clancy2012multispectral}. }

\noindent\textbf{Analytical method to estimate $StO_2$.} \textcolor{black}{A well-established linear model based on modified Beer-Lambert law was used in this paper to obtain ground truth $StO_2$. It uses linear regression to estimate the relative concentrations of oxygenated and deoxygenated haemoglobin ($HbO_2$ and $Hb$), and calculates $StO_2$ as the quantity of $HbO_2$ as a fraction of total haemoglobin ($HbO_2 + Hb$), subject to assumptions \cite{Clancy2015}. Experimental validation has also been carried out in our previous \textit{in vivo} uterine transplantation and bowel surgery experiments \cite{Clancy2015}\cite{Clancy2016} as well as by others \cite{mori2014} \cite{Sorg2005}} The coefficient of determination (CoD) \cite{tjur2009coefficients} was used to evaluate the accuracy of the linear regression estimation.
CoD $\leq 0.85$ is set as threshold for linear regression outliers, and related pixels were excluded for training and evaluation. Pixels located in non-tissue regions, insufficiently illuminated area and specular reflections were also excluded.

\begin{figure}[!ht]   
	\begin{subfigure}[b]{0.24\textwidth}
		\centering
		\includegraphics[width=\textwidth]{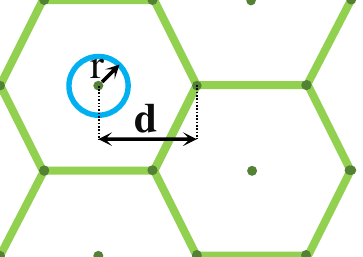}
		\caption{Step1.}
		\label{fig:Mask_step1}
	\end{subfigure}
	\begin{subfigure}[b]{0.24\textwidth}
		\centering
		\includegraphics[width=\textwidth]{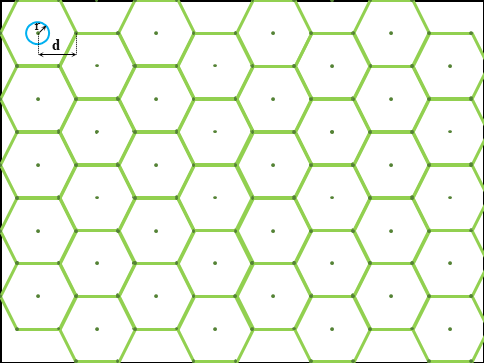}
		\caption{Step2.}
		\label{fig:Mask_step2}
	\end{subfigure}
    	\begin{subfigure}[b]{0.24\textwidth}
		\centering
		\includegraphics[width=\textwidth]{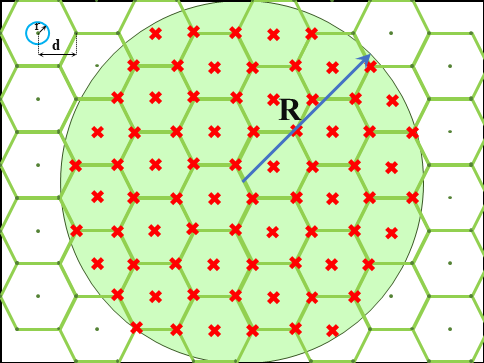}
		\caption{Step3.}
		\label{fig:Mask_step3}
	\end{subfigure}
	\begin{subfigure}[b]{0.24\textwidth}
		\centering
		\includegraphics[width=\textwidth]{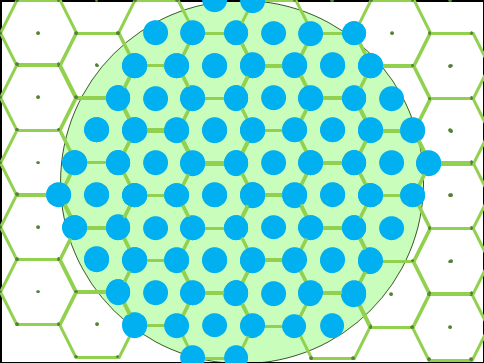}
		\caption{Step4.}
		\label{fig:Mask_step4}
	\end{subfigure}
	\caption{Illustration of the generation of the mask, where fibre core sensing areas (blue circle) are defined with respect to the hexagonal grid in Step 1, and radius of fibre bundle shaded green, individual fibre positions are indicated by red cross in Step 3.} 
    \label{fig:generate_mask}
    \vspace{-1.0em}
\end{figure}

\noindent \textbf{Synthesized sparse hyper-spectral images.}
A number of different distal tip fibre arrangements may be chosen for the experimental hardware. \textcolor{black}{To study how this may affect the performance of the Dual2StO2 network and thereby influence future decisions on the experimental setup, we have simulated data acquired with different fibre arrangements from the dense hyperspectral dataset in Sec.~\ref{sec:Data_And_Preprocessing}.	The use of circular masks with high resolution ground truth images to simulate and assess the performance of imaging fibre bundles has previously been demonstrated \cite{kyrish2010} \cite{shao2018}. Masks were created in MATLAB (R2018a; The Math Works, Inc., USA) using a circular sensing area arranged on a hexagonal grid to represent the array of fibres}, with the spatial information averaged within these areas, as illustrated in Fig.~\ref{fig:generate_mask} and described in the following steps.

\vspace{-1.0em}
\begin{enumerate}[label=Step \arabic*]
	\item \textcolor{black}{Define a radius ($r$, representing the transmissive fibre cores), and horizontal and vertical spacing between the spot centres ($d$, a metric representing the core separations), where the ratio $ \gamma =\frac{r}{d}$ is the fill factor that define the relationship between the area of projected spot and the space. In real situation,  $ \gamma =$ stays unchanged when changing fibre numbers, as the dimensions of individual fibres and their cladding are consistent, which also described in Table.~\ref{table:Result_summary_Dual2StO2_study_v2_equal_gamma}};
    \item \textcolor{black}{Generate a hexagonal grid across the whole image to simulate a hexagonal packed fibre probe ([$W_{s}$,$H_{s}$,$W_{e}$,$H_{e}$] = [$0,0,W,H$], where $W$ and $H$ are the image width and height, the subscripts $s$ and $e$ stands for start and end of the image range respectively)};
	\item \textcolor{black}{Define the radius ($R = \frac{h}{2}$) of the fibre bundle (green circle)}; 
	\item \textcolor{black}{Generate a mask that includes all fibre cores within the bundle};
	\item \textcolor{black}{Finally, average the spatial information within each fibre sensing area to generate a single spectrum for each fibre.}
	\vspace{-1.0em}
\end{enumerate}
\noindent \textbf{Data Augmentation.}
The training data was augmented by horizontal, and vertical flipping, and image cropping using a sliding $96 \times 96$ window with a stride of 16. The cropped \textcolor{black}{images were resized to the target size, $256 \times 256$ through bilinear interpolation, which augments 231 times of 38 original images of and results in 8778 images for training}. In order to maintain information consistency, \textcolor{black}{the central $96 \times 96$ region cropped from original 12 images were resized into a $256 \times 256$ through bilinear interpolation and used as test set}.

\vspace{-2.0em}
\subsection{Dual-input network for StO2 estimation}\label{sec:Dual2StO2}
\vspace{-1.0em}

Dual2StO2 is a cGAN-based image-to-image translation network for $StO_2$ estimation utilizing dual input modalities (RGB, sHSI), which was implemented by Pytorch 4.0.  \textcolor{black}{In analogy with automatic language translation, image-to-image translation defined by Isola \textit{et al.} \cite{pix2pix2016} is a task that translates the representation of one scene into another, which implemented as a general framework called pix2pix for per-pixel classification and regression.  Its fundamental network was based on cGAN, where additional conditions were added for both the generator and discriminator \cite{mirza2014}.}

\begin{figure}[!ht]
	\centering
	\includegraphics[width=0.95\textwidth]{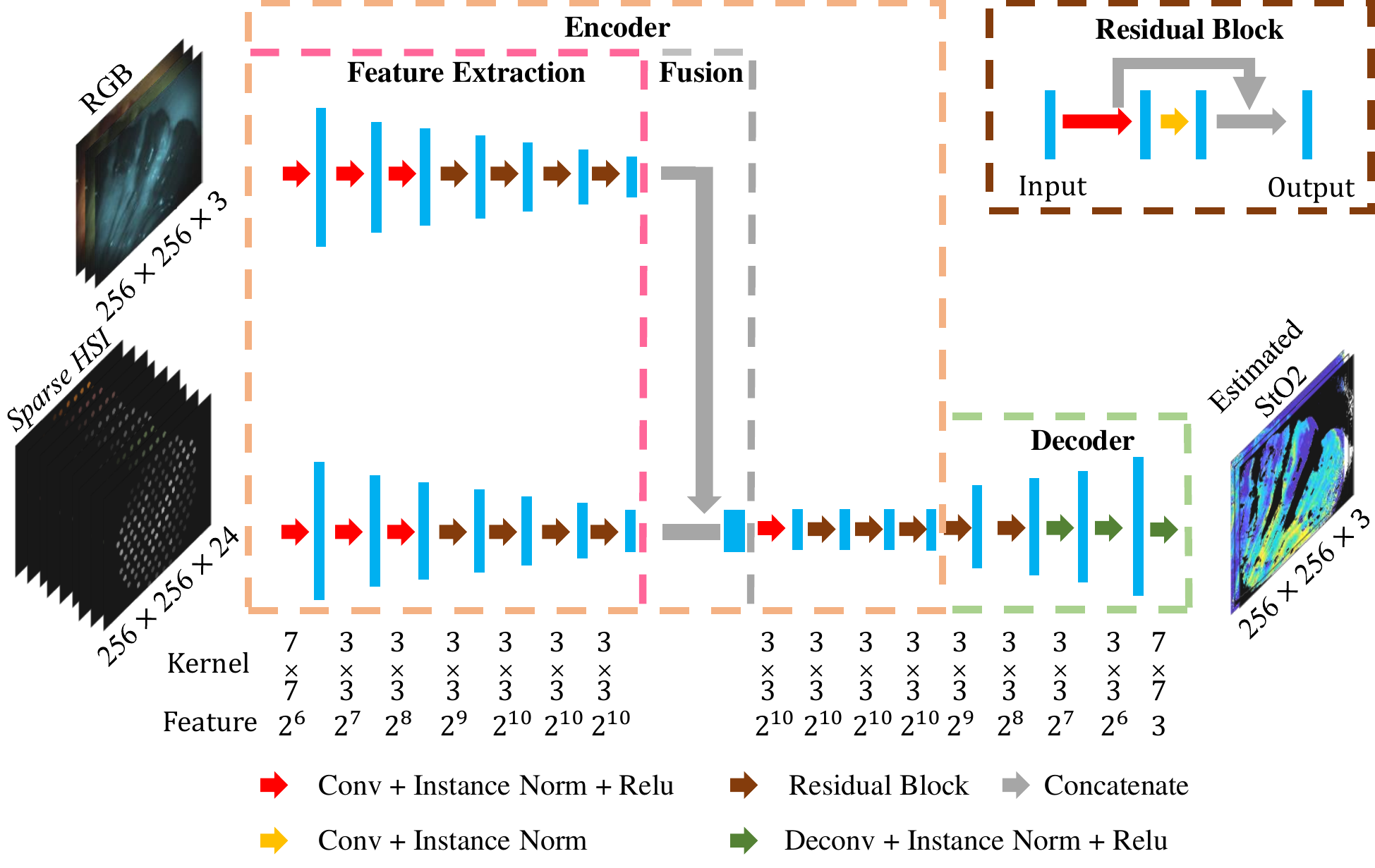}
	\caption{Network architecture of the generator \textbf{G} in proposed Dual2StO2}
	\label{fig:Dual2StO2_generator}
	\vspace{-1.0em}
\end{figure}

\begin{figure}[ht]
	\centering
	\includegraphics[width=0.75\textwidth]{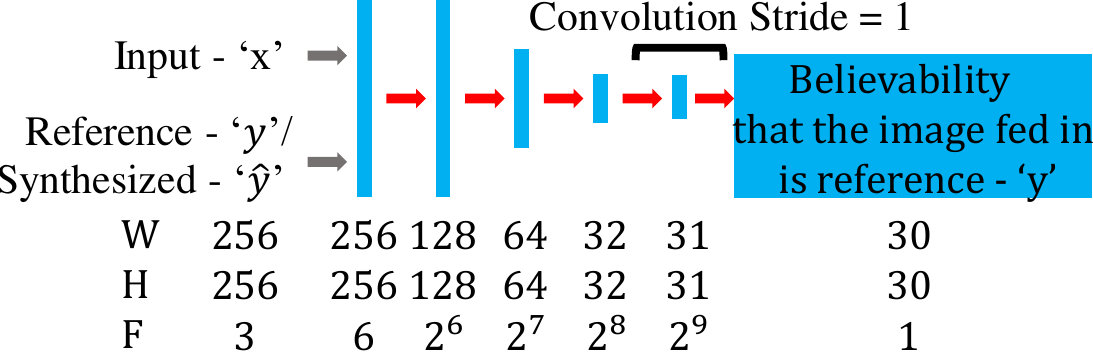}
	\caption{Network architecture of the discriminator \textbf{D} to output believability between 0 to 1 that the image is synthesised $\mathbf{\hat{y}}$ to reference image $y$, where $W$, $H$ and $F$ are the width, height and channel size of feature map.}
	\label{fig:cGAN_network_Discriminator}
    \vspace{-1.0em}
\end{figure}

\noindent \textbf{Generator(G).}
Inspired by the network architecture of pix2pix \cite{pix2pix2016}, \textcolor{black}{it was adopted as the base model in the generator of Dual2StO2, because the relationship between two input modalities (RGB, sHSI) and output ($StO_2$) were known \textcolor{black}{\textit{a priori}} (suitable for supervised learning). The network architecture of the generator was modified based on a multi-input unsupervised learning image-to-image translation framework, called In2I \cite{In2I}, as illustrated in Fig.~\ref{fig:Dual2StO2_generator}}. 
\vspace{-1.0em}
\setlength{\abovecaptionskip}{-0pt}
\begin{itemize}
    \item The encoder (light orange box) was designed to first extract features from the RGB image ($256 \times 256 \times 3$) and sparse HSI ($256 \times 256 \times 24$)  (pink region), fuse the feature map from these two modalities by concatenation (grey box), and extract further features from the fused feature map;
    \item The decoder (light green box) was introduced to \textcolor{black}{decode} the feature map and output the $StO_2$ estimation;
    \item Residual block (brown arrow, with process illustrated in the brown box) proposed by He \textit{et al.} \cite{he2016ResNet} was adopted in both encoder and decoder;
    \item Instance normalization was adopted based on comparison work \cite{ulyanovinstance} \cite{zhou2019normal}, where the results indicated that instance normalization has better performance in image generation tasks than batch normalization;  
    \item One mask was created to filter the position of pixels with saturated pixel values due to specular reflections (NaN), and those with a coefficient of determination (CoD) $\leq 0.85$; 
\end{itemize}
In the training stage of the simulation experiment, the two input modalities (simulated RGB, and the synthesized sparse hyper-spectral image called synthesized sHSI) defined as $S = \{S_{RGB},S_{sHSI}\}$ were \textcolor{black}{fed} into the generator in Fig.~\ref{fig:Dual2StO2_generator} and trained to learn a forward transformation $f_{S\rightarrow T(s)}$ to output a single set of images ($StO_2$) from Sec.~\ref{sec:Data_And_Preprocessing}, under the condition of source domain $S$. Here, the source and target domain were defined by $S$ and $T$, with the data distributions of domain $S$ and $T$ as $p_{data(s)}$ and $p_{data(t)}$. The similar notations in In2I \cite{In2I} are used here.
\noindent  \textbf{Discriminator(D).}
Under the condition of observed image ($\mathbf{x}$) from input domain $S$, the discriminator \textbf{D} will estimate the probability of whether the image is the ground truth image ($\mathbf{y}$) from target domain $T$, or the synthesized image ($f_{S\rightarrow T(s)}, \mathbf{\hat{y}}$) generated by generator \textbf{G}. \textcolor{black}{A convolutional network, called “PatchGAN”, was first introduced by Li \textit{et al.} \cite{li2016} to classify real or fake images based on individual image patches}. A comparison on different patch size was carried by Isola \textit{et al.} \cite{pix2pix2016} and showed that the performance of image-to-image translation were best with $70 \times 70$ patch size. This size was adopted into the implementation of the discriminator.
Concat $(\mathbf{x},\mathbf{y})$ and Concat $(\mathbf{x}, \mathbf{\hat{y}})$ are put into a discriminator separately, which outputs the probability of the input to be $\mathbf{y}$. Here Concat() is concatenate, the probability map is a $30 \times 30 \times 1$ map which is useful for pixel-level rather than image-level translation. The discriminator network architecture is shown in Fig.~\ref{fig:cGAN_network_Discriminator}.
\vspace{-2.0em}
\noindent \textbf{Adversarial learning.}
During the training stage, the generator tries to generate a synthesised image ($f_{S\rightarrow T(s)}$) as real as possible to cheat the discriminator to consider it as real. On the other hand, the discriminator will also improve its ability to make correct judgement on whether the images are the ground truth image from target domain $T$, or the synthesized image ($f_{S\rightarrow T(s)}$) generated by generator. Hence, this forward transformation ($f_{S\rightarrow T(s)}$) is trained by adversarial loss fhe unction Eq.~\ref{eq:Dual2StO2_cGAN}: 
\begin{align} \label{eq:Dual2StO2_cGAN}
    \mathcal{L}_{total} = & \mathcal{L}_{cGAN, S\rightarrow T} + \beta \mathcal{L}_{1}(S\rightarrow T) \nonumber\\ 
    =  & E_{t \sim p_{data(t)}}[\log{D_{T}(t)}] + E_{s \sim p_{data(s)}}[1 - \log{D_{T}(f_{S\rightarrow T(s)})}] \nonumber \\
       & + \beta  E_{s \sim p_{data(s)}}[||(f_{S\rightarrow T(s)}) - t||_{1}] 
    \vspace{-1.0em}   
\end{align}
where D is the discriminator and $\beta$ is the weight of the L1 norm, set as 400 based on previous work \cite{li2018estimation}. 

A minimax two-player game is introduced in this network to train the generator and discriminator through an adversarial process. Hence, the generator in Fig.~\ref{fig:Dual2StO2_generator} is trained to maximize the probability of discriminator to be a false positive. Our final objective is defined by Eq.~\ref{eq:Dual2StO2_cGAN_objective}:
\vspace{-1.0em} 
\begin{equation} \label{eq:Dual2StO2_cGAN_objective}
    \mathcal{G} = \mathop{\arg}\mathop{\min}_{G} \mathop{\max}_{D}[\mathcal{L}_{cGAN}(G,D) + \beta \mathcal{L}_{1}(G)]  
    \vspace{-1.0em} 
\end{equation}
where $G$ is the generator.
\vspace{-1.0em}

\section{Experiments} \label{sec:exp_and_result}
\vspace{-1.0em}
In this section, the evaluation metrics are firstly defined in the Sec.~\ref{sec:exp_metrics} to quantitatively analyse the performance of $StO_2$ estimation, followed by the experiment setup in the Sec.~\ref{sec:exp_setup}.
\vspace{-2.0em}
\subsection{Evaluation metrics.}\label{sec:exp_metrics}
\vspace{-1.0em}
\begin{itemize}
\item \textit{Structural similarity index (SSIM)}:
a perception-based method proposed by Wang \textit{et al.}. \cite{wang2004} comparing the local patterns of pixel intensities that have been normalized for luminance and contrast. \textcolor{black}{Tlshe similarity between the ground truth and synthesized image could be measured and quantified between $0$ and $1$, where $SSIM = 1$ was considered as identical}.
\item \textit{Mean prediction error ($\bar{e}$)}:
the difference in $StO_2$ value between the ground truth and synthesized image at each pixel measured by the L1 norm.
\begin{equation} \label{eq:L1_norm}
    \bar{e}(i,j) = \frac{\sum_{i = 0} ^{W}\sum_{j = 0} ^{H}{e}(i,j)}{n_{effective}}=\frac{\sum_{i = 0} ^{W}\sum_{j = 0} ^{H}||I_{syn{(i,j)}}-I_{gt{(i,j)}}||_{1}}{n_{effective}}
\end{equation}
where $I_{syn}$ and $I_{gt}$ are the absolute values for a pixel at column $i$, row $j$, in the synthesized and ground truth images with width $W$ and height $H$, respectively, and $n_{effective}$ is the total number of pixels in the image excluding saturated and low CoD pixels
\item \textit{Fraction of pixels with high accuracy level ($p_{HAP}$)}: accuracy is defined above a certain level compared to the pixel data in the ground truth image. 
\vspace{-1.0em} 
\begin{equation} \label{eq:HAP}
p_{HAP} = \frac{n_{HAP}}{n_{effective}}
\end{equation}
where $n_{HAP}$ is the number of pixels with high prediction accuracy (i.e. $1-e_{(i,j)} \geq  95\%$), . 
\end{itemize}

\vspace{-3.0em}
\subsection{Experimental set-up.}\label{sec:exp_setup}
\vspace{-1.0em}
\textcolor{black}{Animal} studies were carried out to validate the performance of Dual2StO2 on the \textit{in vivo} acquisitions by separating \textcolor{black}{animal}s into training and test data sets. The training set consisted of 38 acquisitions captured for 10 \textcolor{black}{animal}s (\textcolor{black}{animal} ID: 1 to 10), while 12 further test acquisitions were from the 5 remaining \textcolor{black}{animal}s (\textcolor{black}{animal} ID: 11 to 15).
\vspace{-2.0em}
\setlength{\belowcaptionskip}{0pt}
\begin{figure}[!hb]
	\begin{subfigure}[!h]{0.32\textwidth}
		\centering
		\includegraphics[width=\textwidth]{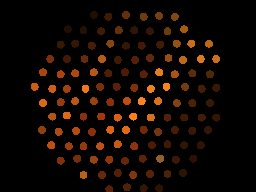}
		\caption{$n_{spot} = 121$}
		\label{fig:sHSI_121}
	\end{subfigure}	
    \begin{subfigure}[!h]{0.32\textwidth}
		\centering
		\includegraphics[width=\textwidth]{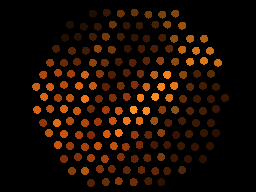}
		\caption{$n_{spot} = 171$}
		\label{fig:sHSI_171}
	\end{subfigure}
	\begin{subfigure}[!h]{0.32\textwidth}
		\centering
		\includegraphics[width=\textwidth]{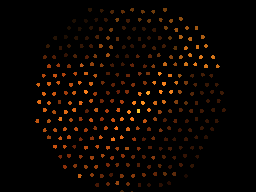}
		\caption{$n_{spot} = 300$}
		\label{fig:sHSI_300}
	\end{subfigure}
	\caption{Synthesized sHSI displayed as three-channel images generated by taking intensities at selected wavelengths ($\lambda = 460, 520, 590 \,\mathrm{nm}$).}
	\label{fig:sHSI_setting}
	\vspace{-2.0em}
\end{figure}
\setlength{\belowcaptionskip}{0pt}

The bundles with circular distal cross sections and different numbers of fibres ($n_{spot}= 0, 121, 171, 300$) were simulated. Two of these configurations (121 and 171 spots) were chosen as they match the existing hardware available, complemented by a fibre bundle with a high number of spots ($300$). To confirm any benefit of integrating sparse HSI, a fibre bundle with zero spots was used as the control group. Fig.~\ref{fig:sHSI_setting} illustrates sample synthesized sHSI images generated by the corresponding masks. These sHSI and simulated RGB images were fed into Dual2StO2. Route 1 in Fig.~\ref{fig:Route} based on the SSRNet developed by Lin \textit{et al.}, was adopted as the baseline to compare the performance of $StO_2$ based on the same simulated RGB and synthesized sHSI as input.
\vspace{-2.0em}
\section{Results.}\label{sec:exp_resulst_analysis}
\begin{figure}[ht]  
    \centering
    \begin{subfigure}[b]{0.49\textwidth}
        \centering
        \includegraphics[width=\textwidth]{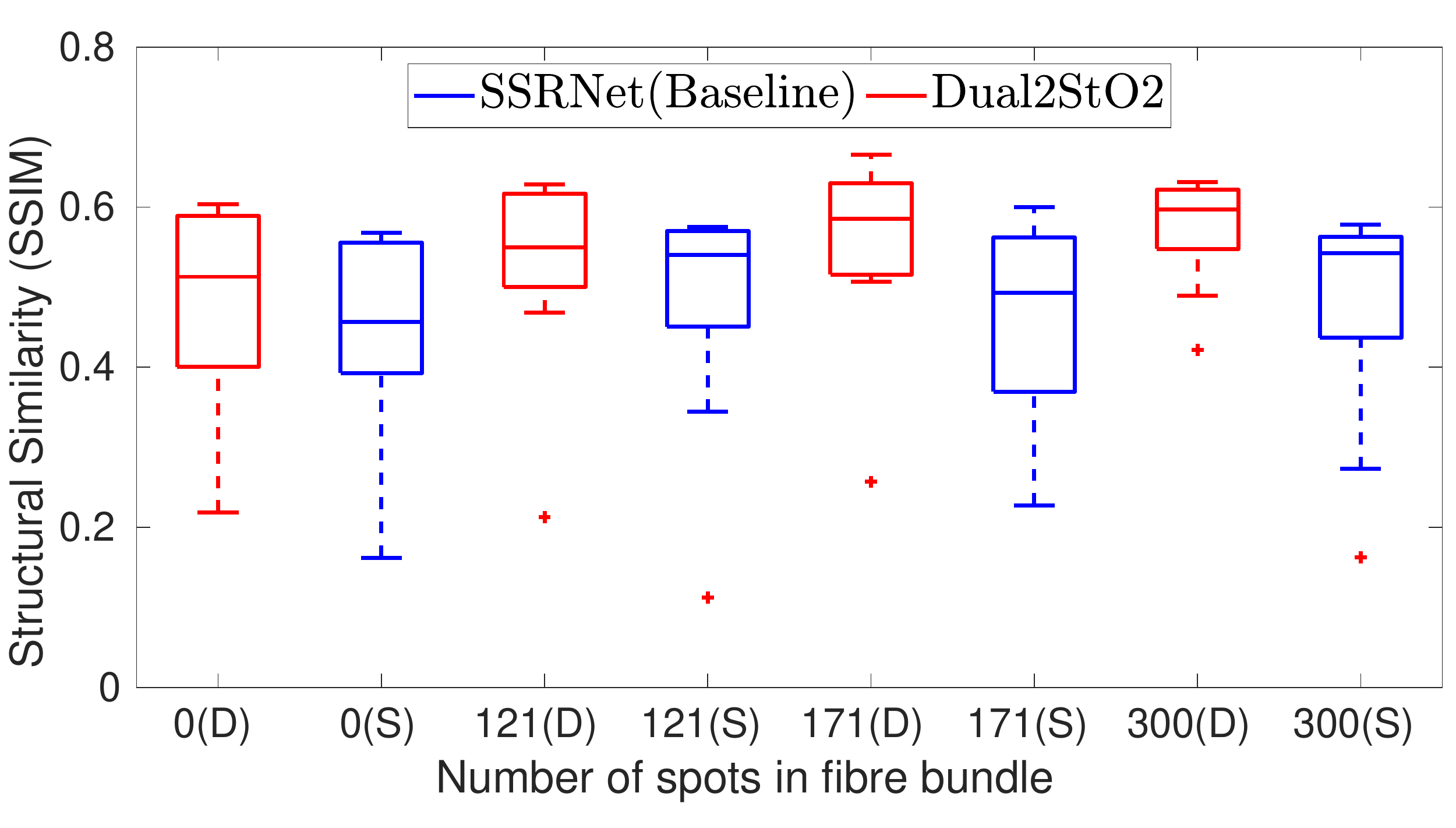}
    	\caption{SSIM.} 
        \label{fig:result_Dual2StO2_Boxplot_SSIM}
	\end{subfigure}
	\hfill
    \begin{subfigure}[b]{0.49\textwidth}
		\centering
    	\includegraphics[width=\textwidth]{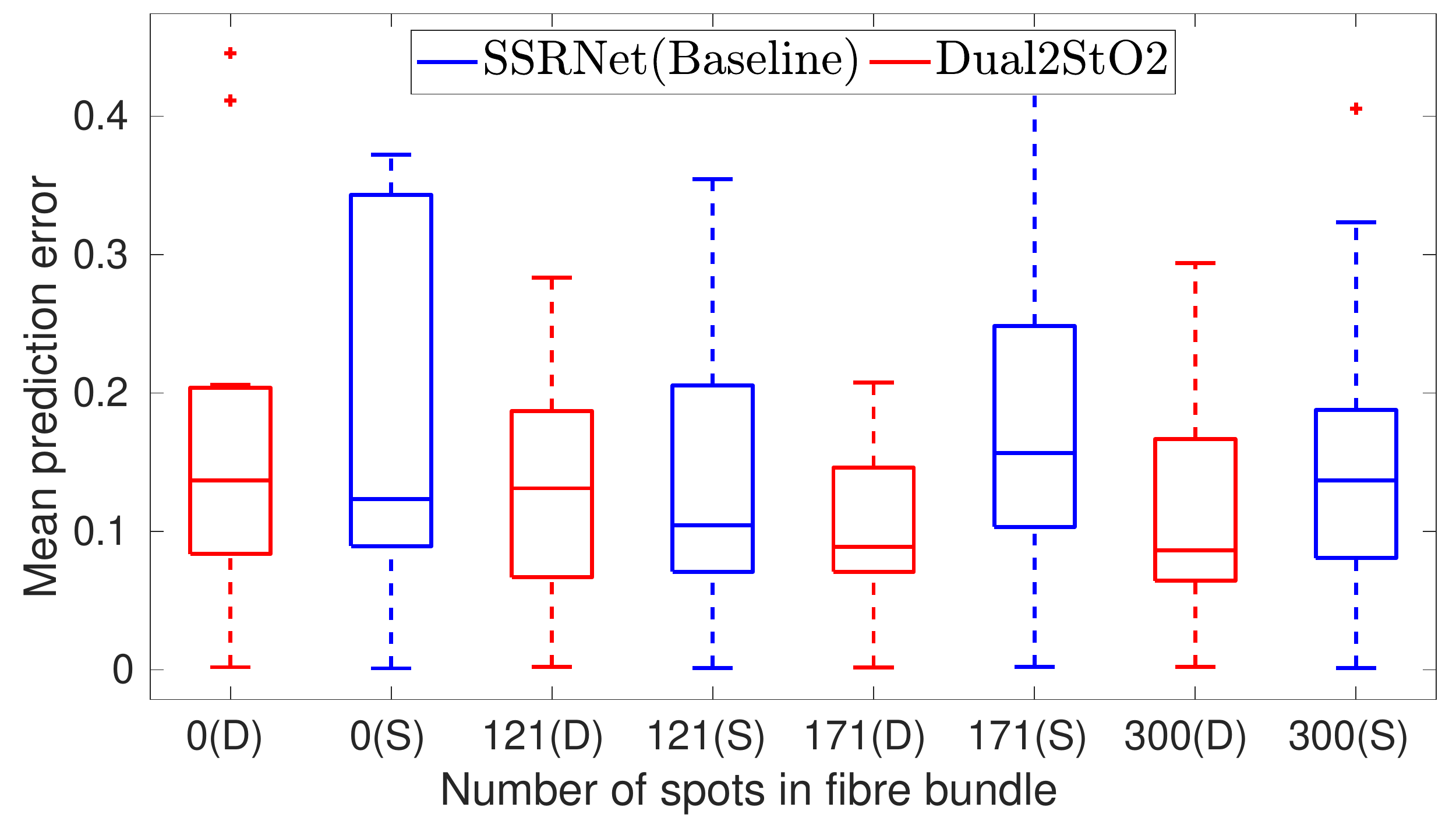}
    	\caption{$\bar{e}$.}
    	\label{fig:result_Dual2StO2_Boxplot_L1}
	\end{subfigure}
	\hfill
	\begin{subfigure}[ht]{0.49\textwidth} 
        \centering
        \includegraphics[width=\textwidth]{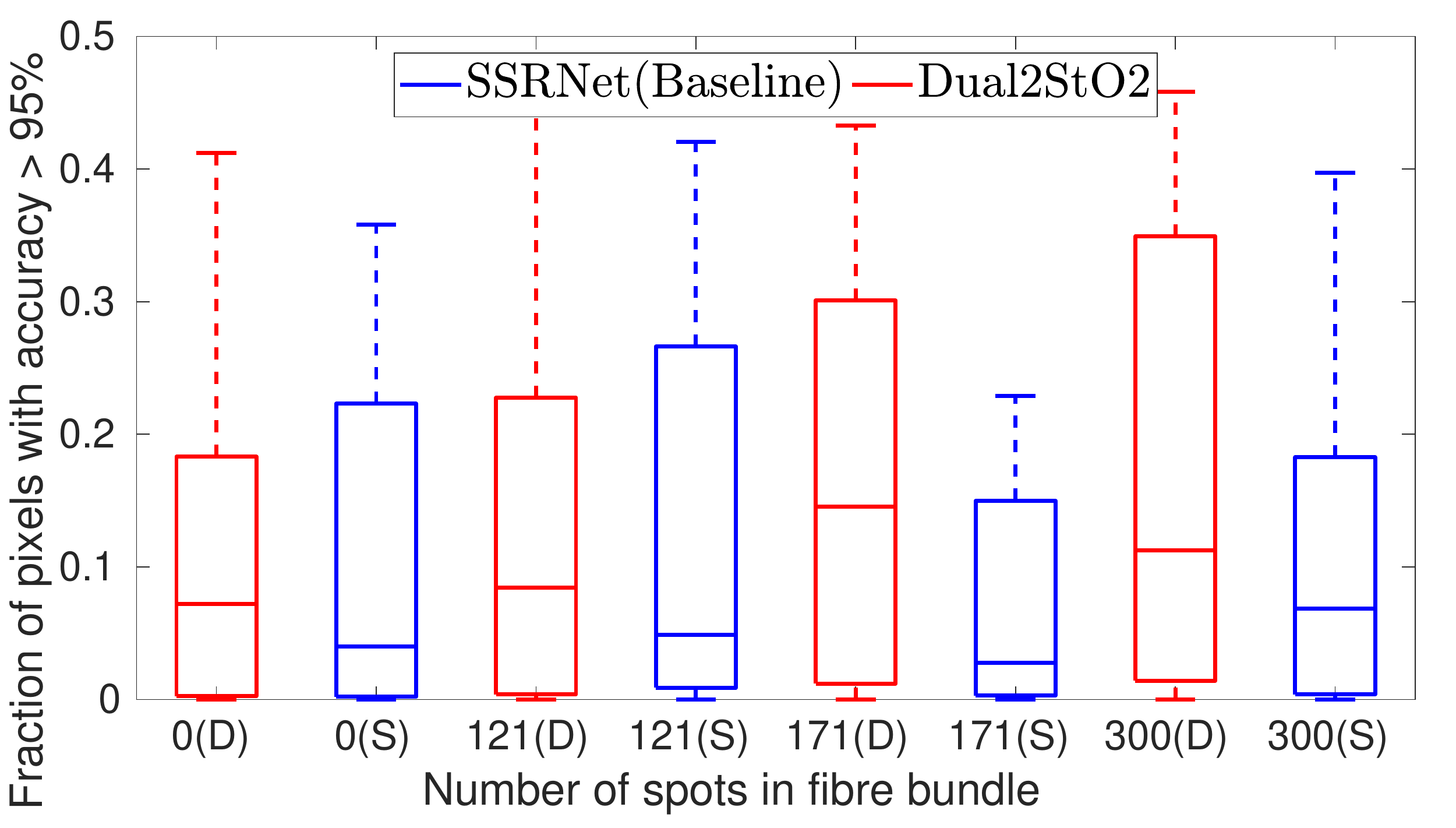}
    	\caption{$p_{HAP}$.} 
        \label{fig:result_Dual2StO2_Boxplot_HCP}
    \end{subfigure}
	\caption{Boxplot of averaged SSIM, averaged mean prediction error ($\bar{e}$) and $p_{HAP}$ in different simulation experiments.}
    \label{fig:result_Dual2StO2_Bar_SSIM_HAP}
    \vspace{-2.0em}
\end{figure}

\vspace{-1.0em}
\textcolor{black}{Table.~\ref{table:Result_summary_Dual2StO2_study_v2_equal_gamma} summarises the performance of Dual2StO2 and SSRNet compared to the ground truth. The proposed network is superior to SSRNet in terms of SSIM and pixel-level accuracy across all fibre bundle configurations. When the fill factor ($\gamma$) is unchanged, the predicted images by Dual2StO2 are structurally closer to the ground truth ($16.5\%$ higher average SSIM), and have $3.6\%$ lower average $\bar{e}$ than SSRNet for $n_{spot} = 300$. Fig.~\ref{fig:result_Dual2StO2_Boxplot_SSIM} shows that even when the number of fibres increased in the bundle the Dual2StO2 predicted images are still structurally closer to the ground truth, with higher SSIM and less variance across different animals indicated by smaller interquartile range (IQR) at high SSIM value than SSRNet. Fig.~\ref{fig:result_Dual2StO2_Boxplot_L1} and Fig.~\ref{fig:result_Dual2StO2_Boxplot_HCP} also presented lower $\bar{e}$ and larger, $p_{HAP}$ by Dual2StO2 than that by SSRNet. Faster $StO_2$ estimation ($\approx35\,\mathrm{ms}$) can be achieved by Dual2StO2 due to its the end-to-end estimation without the intermediate spectral estimation step and light-weight architecture, while the estimation required over $500\,\mathrm{ms}$ by SSRNet \cite{lin2018}. This was validated on a PC (OS: Ubuntu 16.04; processor: i7-3770; graphics card: NVIDIA GTX TITAN X).}
\vspace{-2.0em}
\begin{table}[ht]
\centering
\caption{The average SSIM and average mean prediction error of $StO_2$ with standard deviation estimation by Dual2StO2, SSRNet, and single input network with different sHSI parameters.}
\begin{tabular}{cccccccc}
\hline
\multicolumn{5}{c}{Fibre bundle Setting}                          & Network   & \multicolumn{2}{c}{Results}              \\
$n_{spot}$          &  $\gamma$             &  $r$  &  $d$    & Demo figure       &           & Average SSIM            & Average $\bar{e}$ \\ \hline
\multirow{3}{*}{300} & \multirow{3}{*}{0.25} & \multirow{3}{*}{2.6} &\multirow{3}{*}{10} & \multirow{3}{*}{Fig.~\ref{fig:sHSI_300}} & Dual2StO2 & $\mathbf{0.63 \pm 0.17}$          & $\mathbf{0.11 \pm 0.09}$                \\
                     &                         & & &                    & SSRNet    & $0.54 \pm 0.26$          & $0.15 \pm 0.12$                \\                     
                     &                         & & &                    & \textcolor{black}{Only sHSI}    & \textcolor{black}{$0.54 \pm 0.22$}          & \textcolor{black}{$0.13 \pm 0.11$ }           \\                     
\multirow{3}{*}{171} & \multirow{3}{*}{0.25} & \multirow{3}{*}{3.5} &\multirow{3}{*}{14} & \multirow{3}{*}{Fig.~\ref{fig:sHSI_171}} & Dual2StO2 & $0.61 \pm 0.19$ &  $0.12 \pm 0.1$       \\
                     &                         & & &                    & SSRNet    & $0.52 \pm 0.24$          & $0.17 \pm 0.12$              \\                
                     &                         & & &                    & \textcolor{black}{Only sHSI}    & \textcolor{black}{$0.54 \pm 0.22$}          & \textcolor{black}{$0.15 \pm 0.10$}              \\                     
\multirow{3}{*}{121} & \multirow{3}{*}{0.25} & \multirow{3}{*}{4} &\multirow{3}{*}{16} & \multirow{3}{*}{Fig.~\ref{fig:sHSI_121}} & Dual2StO2 &  $ 0.59 \pm 0.21$         &  $0.14 \pm 0.13$                \\
                     &                         & & &                    & SSRNet    & $0.54 \pm 0.24$         & $0.14 \pm 0.10$              \\
                     &                         & & &                    & \textcolor{black}{Only sHSI}    & \textcolor{black}{$0.53 \pm 0.23$}         & \textcolor{black}{$0.16 \pm 0.10$}              \\                    
\multirow{2}{*}{0}   & \multirow{2}{*}{0}      & \multirow{2}{*}{0} &\multirow{2}{*}{\textbackslash{}}   & \multirow{2}{*}{\textbackslash{}} & Dual2StO2 & $0.53 \pm 0.23$         & $0.17 \pm 0.14$               \\
                     &                         & & &                   & SSRNet    & $0.51 \pm 0.24$         & $0.18 \pm 0.14$                \\
 \hline
\end{tabular}
\label{table:Result_summary_Dual2StO2_study_v2_equal_gamma}
\vspace{-2.0em}
\end{table}
\setlength{\belowcaptionskip}{0pt}
\begin{figure}[!ht]
	\centering
	\includegraphics[clip, trim=2.7cm 5cm 3.cm 10.1cm, width=\textwidth]{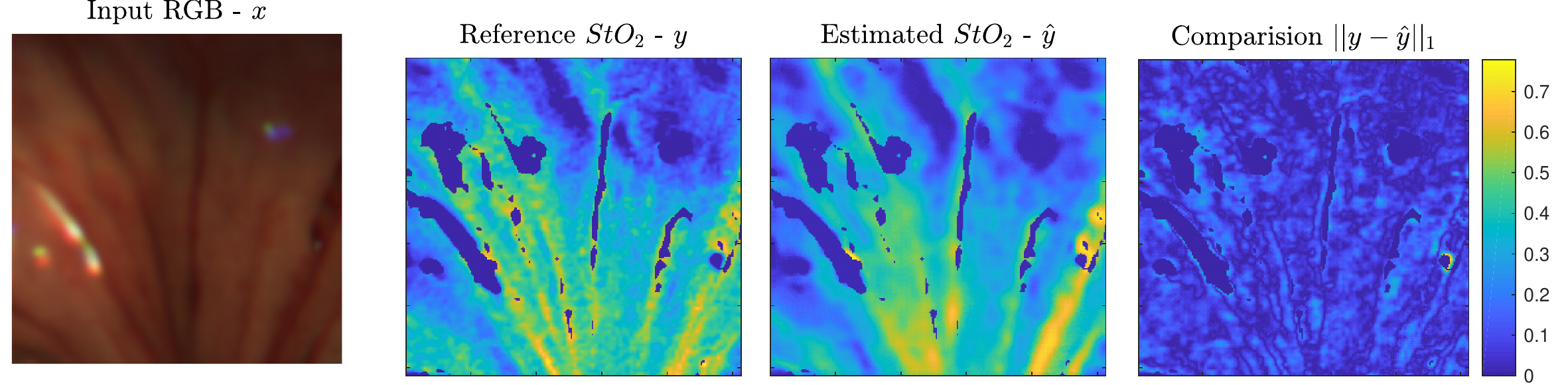}
	\caption{Estimation of $StO_2$ based on Dual2StO2 with $n_{spot} = 300$, which achieved 0.61 in SSIM and 0.07 in $\bar{e}$  compared with the reference $StO_2$.}
	\label{fig:demo_Dual2StO2_good_0}
	\vspace{-2.0em}
\end{figure}
\setlength{\belowcaptionskip}{-10pt}
\begin{figure}[!ht]
	\centering
	\includegraphics[clip, trim=2.7cm 5.cm 3.cm 10.1cm, width=\textwidth]{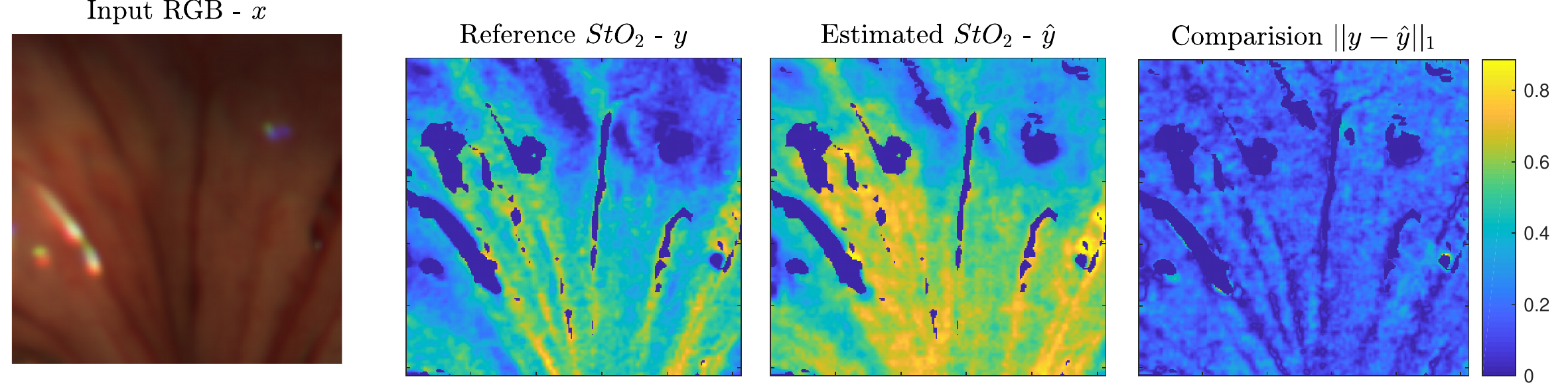}
	\caption{Estimation of $StO_2$ based on SSRNet with $n_{spot} = 300$, which achieved 0.58 in SSIM and 0.15 in $\bar{e}$ compared with the reference $StO_2$.}
	\label{fig:demo_SSRNet_good_0}
	\vspace{-2.0em}
\end{figure}

\textcolor{black}{A better $StO_2$ estimation was achieved with a higher number of fibres in the bundle with $n_{spot} = 300$ achieving the best result for both Dual2StO2 and SSRNet. The overall performance of $StO_2$ estimation was better with additional sHSI information than when compared to that from RGB images only. When sHSI was added, comparing $n_{spot}=121$ to $n_{spot}=0$, the structural similarity increased by $10\%$ and the average mean error reduced by $2.3\%$. An experiment has also been carried out to estimate $StO_2$ using only sHSI. The results from Table.~\ref{table:Result_summary_Dual2StO2_study_v2_equal_gamma} indicate that the single input network can estimate $StO_2$ and achieve  pixel-level accuracy, evaluated by averaged mean prediction error, to some extent. However, the general structure similarity, evaluated by SSIM, is lower than that the dual input network combined with RGB images. As the number of spots increased the performance of $StO_2$ estimation by single input network also improved.}

Fig.~\ref{fig:demo_Dual2StO2_good_0} and Fig.~\ref{fig:demo_SSRNet_good_0} illustrate the typical performance of $StO_2$ estimation by Dual2StO2 and SSRNet with the number of fibres in the bundle ($n_{spot} = 300$) on the second acquisition in \textcolor{black}{animal} ID 13 (porcine bowel). The input RGB image, reference $StO_2$ and estimated $StO_2$ by Dual2StO2 and SSRNet are displayed, while the $StO_2$ value difference between them is also presented. These demonstrate that, with an end-to-end learning training/testing architecture, Dual2StO2 outperforms the two-stage method, i.e. estimating $StO_2$ from hypercubes generated by SSRNet.
\vspace{-3.0em}
\section{Discussion and conclusions}
\vspace{-1.0em}
A dual input network, called Dual2StO2, was designed to estimate $StO_2$ based on sHSI and RGB images. Simulations of three fibre bundles ($n_{spot} = 121$, $171$, $300$) and a control group ($n_{spot} = 0$) were carried out to investigate the impact of integrating sHSI, and to examine the relationship between the number of fibres and prediction accuracy. The results showed that with same fibre bundle, Dual2StO2 has better performance in $StO_2$ estimation (higher SSIM and lower $\bar{e}$ with smaller IQR, larger $p_{HAP}$ and faster prediction) than SSRNet. \textcolor{black}{Compared with the control group ($n_{spot} = 0$, using RGB data alone), the simulation results showed that the overall performance of $StO_2$ estimation with both Dual2StO2 and SSRNet was improved by adding sHSI. Performance was also improved as the number of fibres increased from $121$ to $300$, in terms of prediction accuracy and structural similarity. The result of the control group also indicated that $StO_2$ can be estimated directly from RGB although with consistently lower accuracy. This is in agreement with our previous works \cite{li2018estimation} \cite{Lin2017}. It was also observed that although RGB data could produce realistic spectral estimation, large errors at individual wavelengths were common. While StO2 estimation may be relatively insensitive to these underlying errors, spectral fidelity will be crucial to solving more subtle diagnostic problems such as the detection of cancer. This will be explored further in our future work.}

\textcolor{black}{For real fibre bundles, the transmission characteristics of each fibre differs and cross-talk between fibres may result in measurement noise. This does not affect the result of the Dual2StO2 versus SSRNet comparison, but will affect the spatial accuracy of the sHSI-only results in Table.~\ref{table:Result_summary_Dual2StO2_study_v2_equal_gamma} although it is unlikely to be significant. Furthermore, the sHSI presented here is simulated from an LCTF-based hyperspectral camera, which has lower spectral resolution (10-20 nm) than the spectrograph used in the real SLHSI system ($\approx5$ nm). Therefore, it is likely that the overall StO2 accuracy would be improved when trained with data from the real SLHSI bundle. Nevertheless, the simulations presented here serve as a useful testbed to allow comparative testing of network performance and to guide future design of an optimised fibre bundle. The network architecture of Dual2StO2 will be further customized for better performance, including exploration of custom-designed networks to extract features from RGB and sHSI images separately. The proposed dual input network could potentially be modified to achieve dual output and generate, for example, narrow band images (NBI). The pyramid architecture of multi-generator and discriminator proposed by Wang et al. \cite{wang2017} could also be adopted to enhance the quality of generation. Our network can be further extended to real-time $StO_2$ imaging based on video-to-video synthesis  \cite{wang2018video}}.  
\vspace{-2.0em}
\begin{acknowledgements}
The authors appreciate the assistance of Xiao-Yun Zhou on the GPU configuration and academic advice, and help from Maria Leiloglou. We would like to thank NVIDIA Corporation for the donation of the Titan X GPU. This work was carried out with support from the CRUK Imperial Centre, Imperial ECMC and the NIHR Imperial BRC.\\
\noindent \textbf{Compliance with ethical standards}\\
\textbf{Conflict of interest} The authors declare that they have no conflict of interest.\\
\textbf{Ethical standards} This article does not contain any studies with human participants. All applicable international, national and/or institutional guidelines for the care and use of animals were followed.
\vspace{-1.0em}
\end{acknowledgements}


\bibliographystyle{spbasic}      
\bibliography{library_project}   

\end{document}